\documentclass[lettersize,journal]{IEEEtran}
\usepackage{amsmath,amsfonts}
\usepackage{algorithmic}
\usepackage{algorithm}
\usepackage{array}
\usepackage[caption=false,font=normalsize,labelfont=sf,textfont=sf]{subfig}
\usepackage{textcomp}
\usepackage{stfloats}
\usepackage{url}
\usepackage{verbatim}
\usepackage{graphicx}
\usepackage{cite}
\usepackage{multirow}
\usepackage{hyperref}
\hypersetup{
    colorlinks=true,
    linkcolor=black,
    filecolor=blue,      
    urlcolor=blue,
    citecolor=black,
}
\begin{document}

\title{SuctionNet-1Billion: A Large-Scale Benchmark for Suction Grasping}

\author{Hanwen Cao, Hao-Shu Fang, Wenhai Liu and Cewu Lu 
\thanks{Manuscript received March 29, 2021. accepted August 19, 2021. This letter was recommended for publication by Associate Editor J. Flaco and Editor T. Asfour upon evaluation of the reviewers' comments. This work is supported in part by the National Key R\&D Program of China, No. 2017YFA0700800, National Natural Science Foundation of China under Grants 61772332, Shanghai Qi Zhi Institute, SHEITC(2018-RGZN-02046) and Baidu Fellowship. (\textit{Corresponding author: Cewu Lu}.)}
\thanks{Hanwen Cao is with the Department of Computer Science, Shanghai Jiao Tong University, Shanghai 200240, China, and also with the Department of Electrical and Computer Engineering, University of California, San Dieg, CA 92093 USA (e-mail: h1cao@ucsd.edu).}%
\thanks{Hao-Shu Fang, Wenhai Liu, and Cewu Lu are with the Department of Computer Science, Shanghai Jiao Tong University, Shanghai 200240, China (e-mail: fhaoshu@gmail.com; sjtu-wenhai@sjtu.edu.cn; lucewu@sjtu.edu.cn).}%
\thanks{This article has supplementary downloadable material available at
https://doi.org/10.1109/LRA.2021.3115406, provided by the authors.}
\thanks{Digital Object Identifier 10.1109/LRA.2021.3115406}
}

\markboth{IEEE ROBOTICS AND AUTOMATION LETTERS, VOL. 6, NO. 4, 2021}{Shell \MakeLowercase{\textit{Cao et al.}}: SuctionNet-1Billion: A Large-Scale Benchmark for Suction Grasping}


\IEEEpubid{\begin{minipage}[t]{\textwidth}\ \\[10pt]
        \centering{2377-3766 © 2021 IEEE. Personal use is permitted, but republication/redistribution requires IEEE permission. \\ 
        See https://www.ieee.org/publications/rights/index.html for more information.}
        \end{minipage}} 

\maketitle

\begin{abstract}
	Suction is an important solution for the long-standing robotic grasping problem. Compared with other kinds of grasping, suction grasping is easier to represent and often more reliable in practice. Though preferred in many scenarios, it is not fully investigated and lacks sufficient training data and evaluation benchmarks. To address that, firstly, we propose a new physical model to analytically evaluate seal formation and wrench resistance of a suction grasping, which are two key aspects of grasp success. Secondly, a two-step methodology is adapted to generate annotations on a large-scale dataset collected in real-world cluttered scenarios. Thirdly, a standard online evaluation system is proposed to evaluate suction poses in continuous operation space, which can benchmark different algorithms fairly without the need of exhaustive labeling. Real-robot experiments are conducted to show that our annotations align well with real world. Meanwhile, we propose a method to predict numerous suction poses from an RGB-D image of a cluttered scene and demonstrate our superiority against several previous methods. Result analyses are further provided to help readers better understand the challenges in this area. Data and source code will be publicly available at \url{www.graspnet.net}.
\end{abstract}


\section{Introduction}
\label{sec:intro}

Suction grasping is widely used in real-world pick-and-place tasks. Due to its simplicity and reliability, it is often preferred than parallel-jaw or multi-finger grasping in many industrial and daily scenes~\cite{zeng2018robotic,dexnet4.0,schwarz2018fast}. Though  effective, it actually draws far less attention from researchers than the other types of grasping~\cite{graspnet,gou2021rgb,dexnet2.0,gpd,pointnetgpd}.

Recently, data-driven approach has made great progress in robotic grasping~\cite{graspnet}. Since learning with real-robot is quite time-consuming,  some researchers turned to build simulated environment to obtain large amount of synthetic data~\cite{dexnet2.0,selfsupervisedsuction}. However, learning from synthetic data faces the problem of visual domain gap,
as analyzed in Section~\ref{sec:domain_gap_analysis}. Others tried to collect real-world data and annotate them manually~\cite{zeng2018robotic}. However, the annotation process requires experience in robotic manipulation and is labor-intensive, especially for dense annotations, causing those datasets to be small in scale.
Besides, although previous work~\cite{dexnet3.0} trained with with single-object scenes can transfer to cluttered scenes to some extent, later work~\cite{dexnet4.0} showed that training on cluttered scenes improved the model (see Section~\ref{sec:performance_different_alg}).

To further boost the development of suction grasping algorithms, a benchmark is needed to i) provide large-scale data with cluttered scenes captured by real-world sensors, ii) have dense annotations which is well aligned with real-robot trials, iii) evaluate suction poses quickly in a unified manner without access to a real robot. This is challenging due to the dilemma mentioned above: domain gap for synthetic data and difficulty to annotate real-world data manually. We address this issue by leveraging the advantage of both sides, which means we collect data using real-world sensors and annotate them by analytic computation with a physical model.

\begin{figure}[t]
	\begin{center}
		\includegraphics[width=0.85\linewidth]{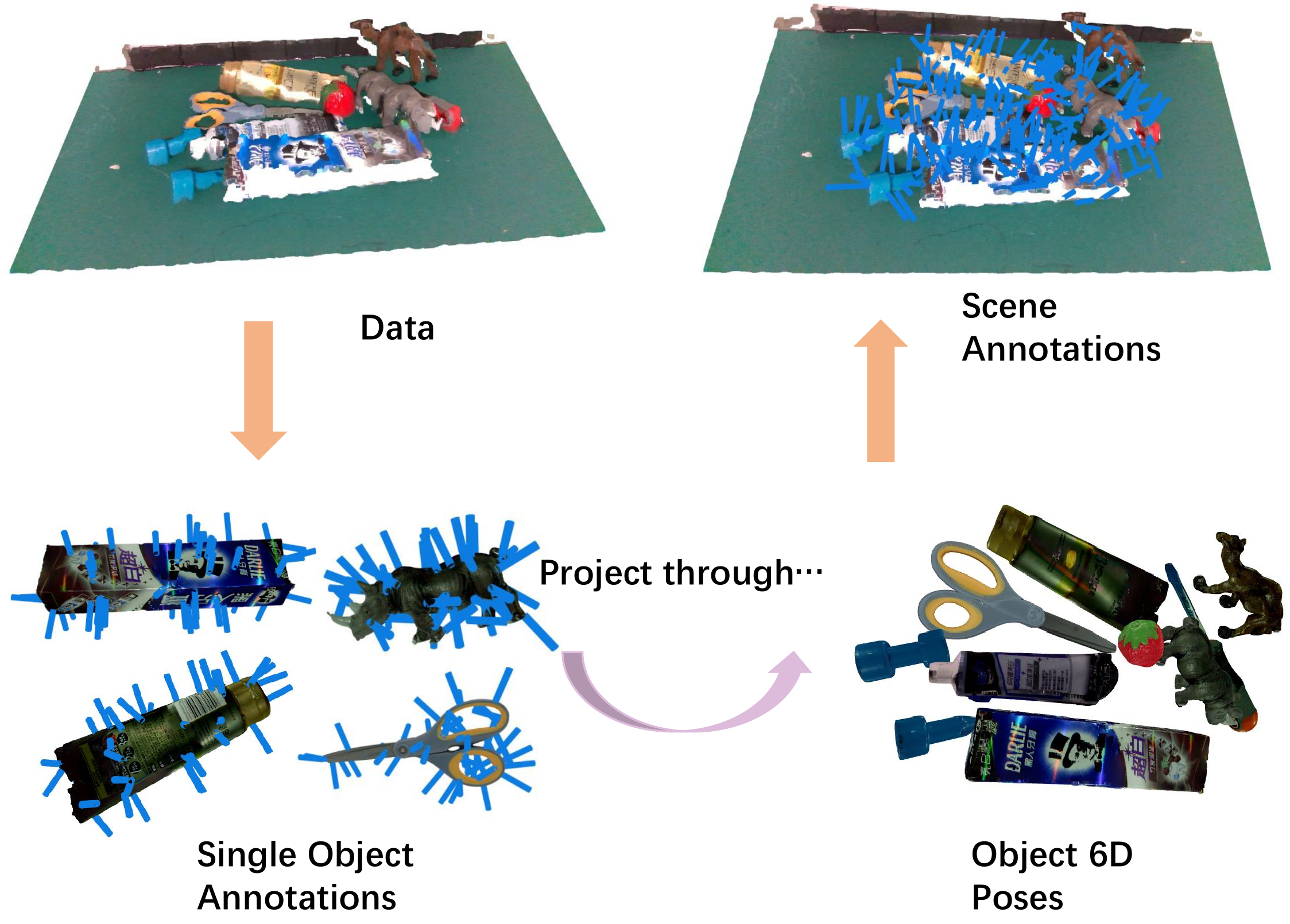} 
	\end{center}
	\vspace{-0.3cm}
	\caption{Our pipeline of building the dataset. For each single object, suction poses are sampled and annotated analytically with a physical model. Scene annotations are obtained by projecting the objects using their 6D poses. The pipeline enables generating large-scale dataset aligning well with real world without intensive human labor.}
	\vspace{-0.5cm}
	\label{fig:teaser}
\end{figure}

To obtain a large-scale densely annotated suction grasping dataset, we get inspiration from previous literature~\cite{graspnet} and propose a two-step annotation pipeline. Specifically, we first generate and annotate suction poses on 88 single objects with their accurate 3D mesh models. Then we utilize objects' 6D poses to generate dense annotations for totally 190 cluttered scenes by projection and collision detection. Each scene contains 512 RGB-D images captured by two cameras from 256 different views, leading to \textbf{97,280} images in total. Contributed to the effective annotation methodology, we built the first large-scale real-world suction grasping dataset in cluttered scenes with more than \textbf{one billion} suction annotations, which is larger than previous datasets by orders of magnitude. In addition, we provide an online evaluation system to quickly evaluate suction grasping algorithms in a unified manner. We conducted real-world experiments to show that our benchmark aligns well with real-robot manipulations. The pipeline of building our dataset is shown in Fig.~\ref{fig:teaser}. We hope the large-scale densely annotated real data as well as the evaluation metric can benefit related research.

With the new large-scale dataset, we propose a new method to predict 6D suction poses directly from single RGB-D image. Firstly, a neural network predicts a pixel-wise seal formation heatmap and a center map for objects from the RGB-D input. Then, we sample high-score suction candidates by grid sampling. Finally, we compute 3D suction points and directions from depth image geometrically with camera intrinsics. The effectiveness of our proposed method is demonstrated by experiments. Previous methods~\cite{dexnet3.0,zeng2018robotic} are also evaluated on our benchmark.

The rest of our paper is organized as follows: we firstly introduce previous methods and datasets in Sec.~\ref{sec:related}. Next, in Sec.~\ref{sec:benchmark}, we detail the construction of our benchmark, including our analytical models, annotation procedure and evaluation metrics. Our network is then introduced in Sec.~\ref{sec:method}. Finally, abundant analyses to our benchmarks and several baseline methods are carried out in Sec.~\ref{sec:exp}.

\section{Related Work}
\label{sec:related}
In this section, we will review some previous suction-based grasping algorithms, previous suction-related datasets and suction analytic models.

\subsection{Suction Datasets}

Zeng \textit{et al.}~\cite{zeng2018robotic} built a real-world dataset by seeking experts who had experience with suction grasp to annotate suctionable and non-suctionable areas on raw RGB-D images. However, the labor-intensive procedure and requirement of expertise makes it hard to build large-scale dataset. In the contrast,  Mahler \textit{et al.}~\cite{dexnet2.0,dexnet3.0} turned to simulation and built a synthetic dataset by using 3D models 
and annotated suctions by analytical computing. The dataset contained depth images rendered from those 3D models. They built a much larger dataset but suffered from the problem of domain gap of visual perception and did not have RGB images. ~\cite{dexnet4.0} also pointed out that methods trained on single object scenes does not transfer well to cluttered scenes. Other datasets~\cite{deft,cartman} are either small in scale or publicly unavailable. 
In this work, we aim to provide a dataset that is both large in scale and aligns with real visual perception. An online evaluation system is further provided to enable algorithm evaluation without the need to access real robots

Meanwhile, our dataset contains 6D poses and thus can be used by researchers in 6D pose estimation community~\cite{kleeberger2019large, posecnn, rennie2016dataset}. And these datasets can be used for suction purpose with our annotation pipeline.

\subsection{Suction Analytic Models}

In the literature of graphics, Provot \textit{et al.}~\cite{deformation} modeled deformable objects like cloth and rubber using a spring-mass system with several types springs. Mahler \textit{et al.}~\cite{dexnet3.0} followed this work and modeled the suction cup as a quasi-static spring system. They evaluated whether the suction can form a seal against object surface by estimating the deformation energy. Readers can refer to a review by Ali \textit{et al.}~\cite{rubberreview} for more constitutive models of rubber-like materials.

Assuming a seal is formed, several models have been designed to check force equilibrium. Most early works regarded the suction cup as rigid and considered forces along surface normals, tangential forces due to friction and pulling forces due to motion. Bahr \textit{et al.}~\cite{bahr1996design} and Morrison \textit{et al.}~\cite{cartman} augmented the model by further considering moment resistance and torsional friction . Mahler \textit{et al.}~\cite{dexnet3.0} combined both torsional friction and moment resistance. We follow the suction model in \cite{dexnet3.0} due to its effectiveness and make some modifications to achieve fast and accurate annotations.

\subsection{Suction-based grasping}

Eppner~\cite{lessons} firstly detected objects and then pushed objects from top or side until suction was achieved. Hernandez \textit{et al.}~\cite{deft} firstly recognized objects and estimated their 6D poses~\cite{he2020pvn3d, zhao2018estimating}. They then generated suction candidates on the 3D model of the item based on primitive shapes. Those candidates were finally scored based on geometric and dynamic constraints. Zeng \textit{et al.}~\cite{zeng2018robotic} used a neural network trained on a human-labeled dataset to predict grasping affordance map from the RGB-D input and chose the suction point with max affordance value. Morrison \textit{et al.}~\cite{cartman} firstly obtained semantic segmentation of the scenes and scored the suctions using heuristics like distance from boundary and object curvature. Mahler \textit{et al.}~\cite{dexnet3.0} designed a network to score a suction candidate with the depth image centered at the suction point and suction pose information as input.
Correa \textit{et al.}~\cite{toppling} further proposed a toppling strategy to expose access to robust suction points and increase the suction reliability. Shao \textit{et al.}~\cite{selfsupervisedsuction} proposed an online self-supervised learning method by predicting suction and getting results from a simulated environment.

\section{SuctionNet-1Billion}
\label{sec:benchmark}
In this section, we provide details of our benchmark and illustrate how we build it.

\subsection{Overview}
Previous suction datasets are either small in scale~\cite{zeng2018robotic} due to labor intensity and expertise required for annotators, or only contain synthetic data~\cite{dexnet3.0,dexnet4.0} which can cause performance decrease in real world due to the domain gap. Also, DexNet3.0~\cite{dexnet3.0} only focuses on single objects, which does not transfer well to cluttered scenes pointed out by~\cite{dexnet4.0}. To address those problems, we build a large-scale real-world dataset with cluttered scenes and dense suction pose annotations named \textit{SuctionNet-1Billion}. 

Our data is from GraspNet-1Billion~\cite{graspnet}, which contains 88 daily objects with high quality 3D mesh models and 190 cluttered scenes. Each scene contains 512 RGB-D images captured by two different cameras, RealSense and Kinect, from 256 views. Accurate object 6D pose annotations, object masks, bounding boxes and each view's camera pose are also provided. For each image, we densely annotate suction poses by computing both seal formation scores and wrench resistance scores. Our dataset contains 3.3M $\sim$ 8.1M annotations for each scene and 1.1B in total. 

\subsection{Suction Pose Annotation}
\label{sec:pose_anno}
We define a suction pose as a 3D suction point and a direction vector starting from the suction point and pointing outside of the object surface. Inspired by DexNet 3.0~\cite{dexnet3.0}, given a suction pose, we evaluate seal formation and wrench resistance separately. However, instead of labeling suction poses with binary scores, we assign them two continuous scores, namely seal score $S_{seal}$ and wrench score $S_{wrench}$, both in the range of $[0, 1]$. We define the overall score as the product of these two scores $S = S_{seal} \times S_{wrench}$. In the following, we will describe how we compute the scores and present the overall annotation procedure.

\begin{figure}[t]
	\begin{center}
		\includegraphics[width=0.8\linewidth]{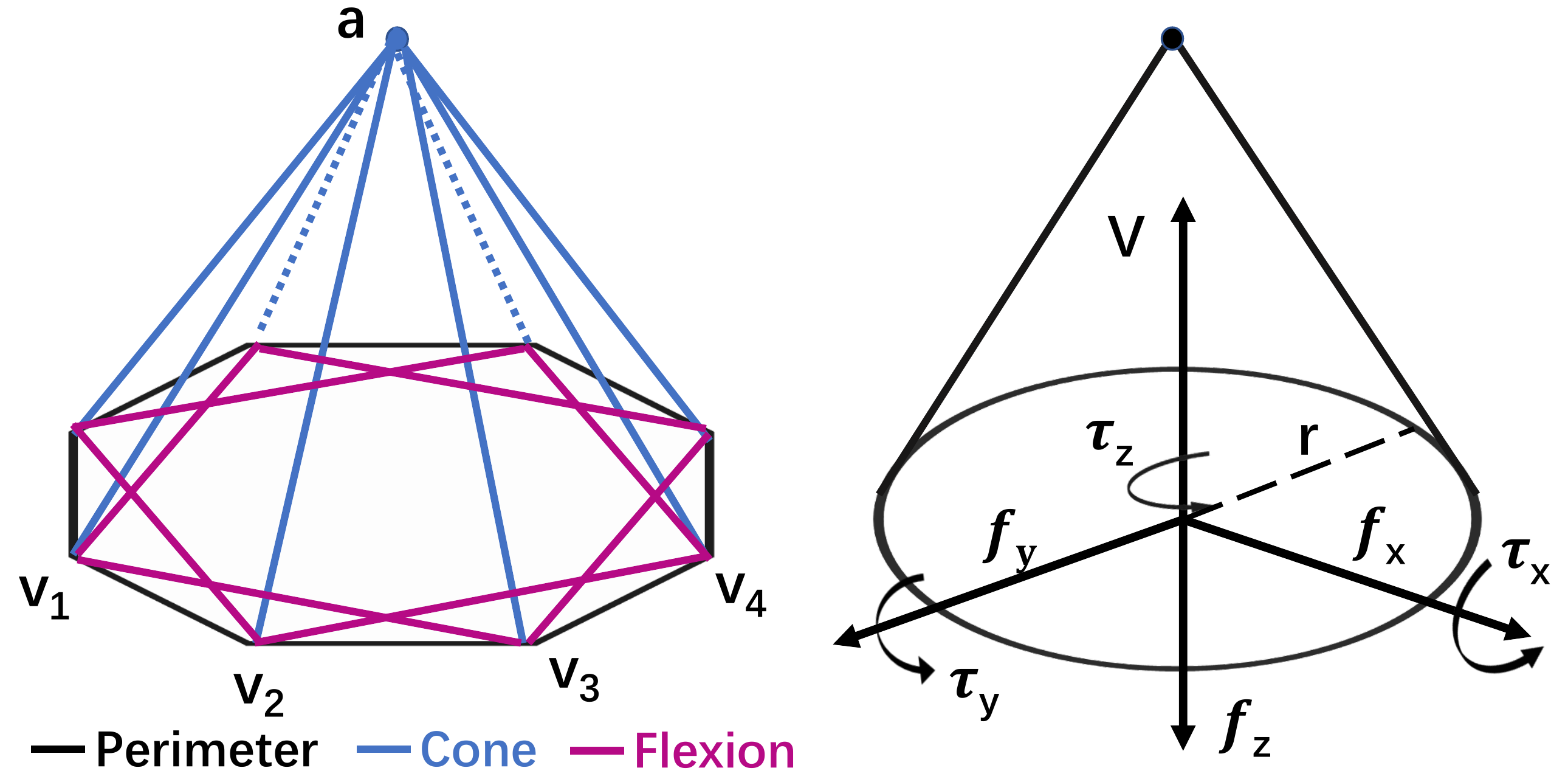} 
	\end{center}
	\vspace{-0.3cm}
	\caption{DexNet 3.0~\cite{dexnet3.0} suction model. \textbf{Left}: the quasi-static spring model with three different springs represented by different colors. \textbf{Right}: Wrench basis for the suction cup. Best viewed in color.}
	\label{fig:dexnet3.0}
	\vspace{-0.4cm}
\end{figure}

\subsubsection{\textbf{Seal Score Evaluation}}

DexNet 3.0~\cite{dexnet3.0} models a suction cup as a quasi-static spring system with topology shown in Fig.~\ref{fig:dexnet3.0} left. Specifically, the model has three kinds of springs: 1) Perimeter Springs -- those linking vertex \textbf{v}$_i$ to vertex \textbf{v}$_{i+1}$; 2) Cone Springs -- those linking vertex \textbf{v}$_i$ to vertex \textbf{a}, ; 3) Flexion Springs -- those linking vertex \textbf{v}$_i$ to vertex \textbf{v}$_{i+2}$. 

They project the cup along the suction direction to the suction point on the object surface. The length of each spring will change due to the irregularity of the surface. If each spring's length change is within $10\%$, seal formation is labeled as positive, otherwise negative.

We modify the annotation process as follows. Firstly, we remove Cone Springs and Flexion Springs and only keep Perimeter Springs since the three types of springs are coupled with each other. 
Secondly, we model the score as a continuous value instead of a binary one. Define $l_i$ as the length of the perimeter spring linking vertex \textbf{v}$_i$ and vertex \textbf{v}$_{i+1}$. Given the original length $l_i$ and the length after projection $l'_i$, we get the change ratio as:
\begin{equation}
\label{eq:change_ratio}
r_i = min(1, \lvert \frac{l'_i-l_i}{l_i} \rvert). 
\end{equation} 
The spring deformation score is then defined as:
\begin{equation}
\label{eq:deform_score}
S_{deform} = 1 - max\{r_1, r_2, ..., r_n\}.
\end{equation} 
Thirdly, to compensate for accuracy loss induced by representing a circle as polygon, we fit a plane for object surface points around the contact ring using least-squares solution and calculate the average of squared errors from the surface points to the plane, denoted as $E_{square}$. 
The plane fit score is defined as:

\begin{equation}
\label{eq:fit_score}
S_{fit} = e^{-c \cdot E_{square}}.
\end{equation}

Where $c>0$ is some fixed coefficient. Finally, the seal score is defined as the product of $S_{deform}$ and $S_{fit}$:

\begin{equation}
\label{eq:seal_score}
S_{seal} = S_{deform} \times S_{fit}.
\end{equation}
Note that we check if all the vertexes of suction gripper can be projected onto the object before the computation. If any vertex falls out of the object (\textit{e.g.} suction center is near the object edge), $S_{seal}$ will be 0.

We conduct real-robot experiments in Sec.~\ref{sec:realrobot} to demonstrate that our annotations align well with real world suctions.

\begin{figure}[ht]
	\begin{center}
		\includegraphics[width=0.95\linewidth]{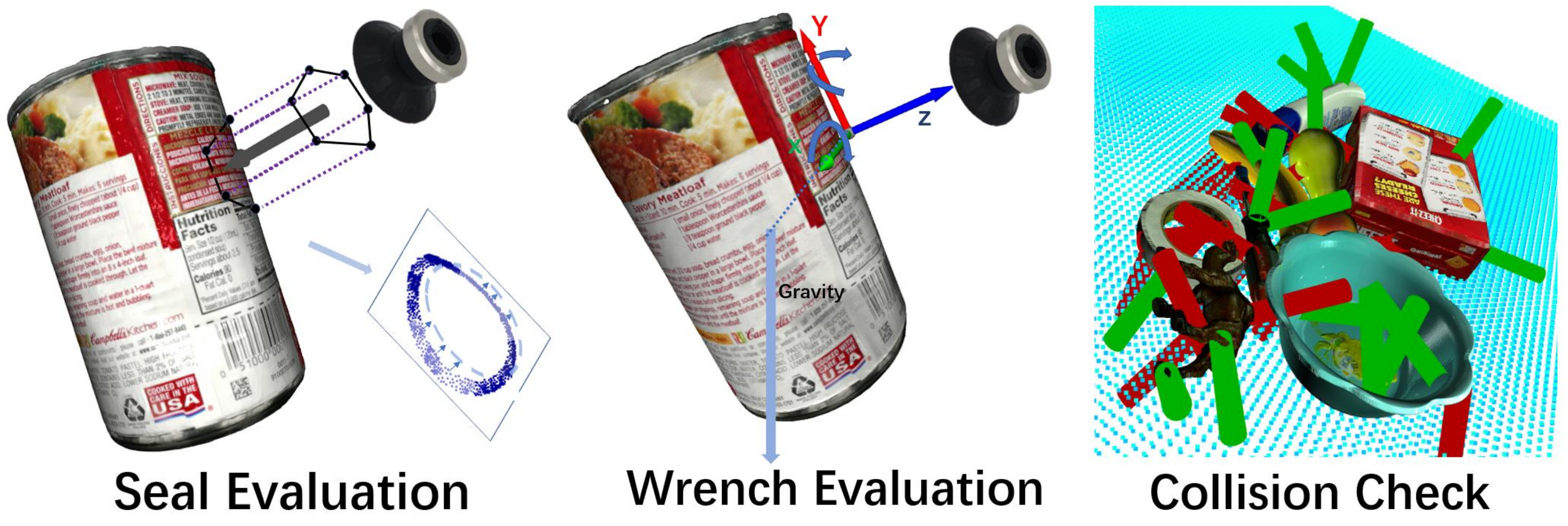} 
	\end{center}
	\vspace{-0.3cm}
	\caption{Our annotation Procedure. \textbf{Left}: For each object, Seal Formation Score $S_{seal}$ is annotated by projecting the suction cup to the object; then max deformation ratio of the springs and the average residual errors from the points around the contact ring and a fitted plane are computed to get the score. \textbf{Middle}: Wrench Score $S_{wrench}$ is computed after projecting the objects with their 6D poses in a scene considering the wrench caused by gravity around axis X and Y. \textbf{Right}: Scene collision is checked in each scene where red represents collision while green represents collision-free. Best viewed in color.}
	\vspace{-0.2cm}
	\label{fig:annotation}
\end{figure}

\subsubsection{Wrench Score Evaluation}

To lift up an object, the suction end-effector should be able to resist the wrench caused by gravity. DexNet 3.0~\cite{dexnet3.0} defined 5 kinds of basic wrenches, namely 1) Actuated Normal Force ($f_z$); 2) Vacuum Force ($V$); 3) Frictional Force ($f_f = (f_x, f_y)$); 4) Torsional Friction ($\tau _z$) and 5) Elastic Restoring Torque ($\tau _e = (\tau _x, \tau _y)$) as shown in Fig.~\ref{fig:dexnet3.0} right. The following 3 constraints are defined:
\begin{equation}
\label{eq:dexnet_wrench}
\begin{split}
Friction:& \sqrt{3} |f_x| \leq \mu f_N \quad \sqrt{3}|f_y| \leq \mu f_N \quad \sqrt{3}|\tau _z| \leq r\mu f_N, \\
Material:& \sqrt{2}|\tau _x| \leq \pi r k \quad \sqrt{2}|\tau _y| \leq \pi r k, \\
Suction:& f_z \geq -V,
\end{split}
\end{equation}
where $\mu$ is the friction coefficient, $f_N = f_z + V$ is the normal force between the suction cup and object, $r$ is the radius of the contact ring and $k$ is a material-related constant modeling the maximum stress for which the suction cup has linear-elastic behavior. They label the wrench resistance as positive if all the constraints are satisfied otherwise negative.

We simplify the wrench score evaluation by removing the Friction and Suction constraints in Eq.~\ref{eq:dexnet_wrench}. The reason we remove the Friction constraint is that it requires the suction cup to stick on the object surface while keeping the seal, which is usually satisfied in suction manipulations.  As for the Suction constraint, it does not contribute to the suction policy because if $V$ is less than the object gravity, then all suctions will fail no matter where the suction point is. Therefore, we assume this constraint is always satisfied.

The wrench score is also modified to be continuous in $[0, 1]$. Specifically, we define a torque threshold as $\tau _{thre} = \pi r k$ and overall tangential torque as $|\tau _e| = \sqrt{\tau_x^2 + \tau_y^2}$. The wrench score is defined as:
\begin{equation}
\label{eq:wrench_score}
S_{wrench} = 1 - min(1, |\tau_e| / \tau_{thre}).
\end{equation}
Note that $|\tau_e|$ is actually related to the weight of object but we simply assign the same weight to each object. 
The weight actually does not contribute to the policy because given any weight $m$, the gravity wrench is $\tau = mgl$ where $g$ is the gravity acceleration constant. The policy should try to minimize the force arm $l$ which is dependent on the suction point and direction (i.e. suction pose) regardless of the weight. In other words, the optimal suction pose is actually the same for different weights. 


\begin{table*}[ht]
	\centering
	\caption{Ground truth score distribution for different splits.}
	\vspace{-0.2cm}
	\label{tab:anno_distribution}
	\resizebox{\textwidth}{15mm}{
		\begin{tabular}{|c|ccc|ccc|ccc|ccc|}
			\hline
			\multirow{2}*{Range} & \multicolumn{3}{c|}{Train} & \multicolumn{3}{c|}{Test-Seen} & \multicolumn{3}{c|}{Test-Similar}       & \multicolumn{3}{c|}{Test-Novel}          \\ 
			\cline{2-13} & $S_{seal}$ & $S_{wrench}$     & $S$ 
			& $S_{seal}$ & $S_{wrench}$     & $S$     & $S_{seal}$ &  $S_{wrench}$    & $S$     & $S_{seal}$ &  $S_{wrench}$    & $S$     \\ 
			\hline
			$[0.0, 0.2)$ & 35.5\%   & 24.8\% & 55.5\% & 35.3\%   & 23.4\% & 53.0\% & 27.4\%   & 18.2\% & 41.8\% & 72.4\%   & 24.5\% & 82.8\% \\
			
			$[0.2, 0.4)$ & 5.5\%   & 11.7\% & 12.0\% & 5.5\%   & 11.5\% & 12.2\% & 3.7\%   & 13.6\% & 13.9\% & 4.5\%   & 13.2\% & 7.0\% \\
			
			$[0.4, 0.6)$ & 5.6\%   & 22.6\% & 17.0\% & 5.5\%   & 22.6\% & 17.2\% & 4.1\%   & 26.5\% & 23.2\% & 4.2\%   & 23.6\% & 6.1\% \\ 
			
			$[0.6, 0.8)$ & 9.4\%   & 26.9\% & 11.9\% & 9.4\% & 28.9\% & 13.7\% & 9.6\% & 28.5\% & 16.0\% & 4.7\% & 26.6\% & 3.3\% \\ 
			
			$[0.8, 1.0]$ & 43.9\%   & 13.9\% & 3.6\% & 44.4\% & 13.5\% & 3.8\% & 55.3\% & 13.3\% & 5.2\% & 14.2\% & 12.1\% & 0.8\% \\ \hline
		\end{tabular}
		
	}
	\vspace{-0.3cm}
\end{table*}

\begin{table}[ht]
	\centering\footnotesize
	\caption{Summary of the properties of publicly available suction datasets. "Obj.", "img." and "Modal." are short for Objects, Images and Modality. "Cam." and "Sim." are short for Camera and Simulation. "-" means unknown.}
	\vspace{-0.2cm}
	\label{tab:comparison}
	\begin{tabular}{c|c|c|c|c|c|c|c}
		\hline
		Data    & \begin{tabular}[c]{@{}l@{}}Labels\\ /scene\end{tabular} & \begin{tabular}[c]{@{}l@{}}Obj.\\ /scene\end{tabular} & \begin{tabular}[c]{@{}l@{}}Total \\ labels\end{tabular} & \begin{tabular}[c]{@{}l@{}}Total \\ img.\end{tabular} & Modal. & Data & CAD \\ \hline
		\cite{dexnet3.0} & 1 & 1 & 2.8M & 2.8M & depth & sim. & Yes  \\ 
		\hline
		\cite{dexnet4.0} & 1 & 1 & 2.5M & 2.5M & depth & sim. & Yes  \\ 
		\hline
		\cite{zeng2018robotic}  & $\sim$100K & - & 191M & 1837 & rgbd & 1 cam. & No \\ 
		\hline
		\cite{posecnn}  & None & $\sim$5 & None & 134K & rgbd & 1 cam. & Yes \\ 
		\hline
		\textbf{ours} & \textbf{$\sim$5.1M} & \textbf{$\sim$10}  & \textbf{1.1B} & \textbf{97K} & \textbf{rgbd} & \textbf{2 cam.} & \textbf{Yes} \\ 
		\hline
	\end{tabular}
	\vspace{-0.3cm}
\end{table}

\subsubsection{Overall Procedure}

\paragraph{6D Pose Annotation} We adopt the 6D pose annotations from GraspNet-1Billion~\cite{graspnet}. The annotations are obtained by propagating the annotated 6D poses of the first frame of each scene using the recorded camera poses:
\begin{equation}
\label{eq:pose_anno}
\textbf{P}_i^j = \textbf{cam}_i^{-1} \textbf{cam}_0 \textbf{P}_0^j,
\end{equation}
where $\textbf{P}_i^j$ is the 6D pose of object j at frame i and $\textbf{cam}_i$ is the camera pose of frame $i$. We refer readers to~\cite{graspnet} for more details about object collection and 6d pose annotation.

\paragraph{Suction Pose Annotation} 
Following GraspNet-1Billion~\cite{graspnet}, we adopt the two-step pipeline to annotate suction poses: annotate single objects and then project to scenes. 

Firstly, we sample and annotate suction poses on each single object. The 3D mesh models are first voxelized (voxel side length is 5 $mm$) and suction points with the corresponding directions (surface normals) are sampled uniformly in voxel space.

Secondly, for each scene, these suction poses are projected using the annotated 6D object poses:
\begin{equation}
\label{eq:project}
\begin{split}
&\textbf{P}^i = \textbf{cam}_0 \textbf{P}^i_0, \\
&\mathbb{S}^i_{(w)} = \textbf{P}^i \cdot \mathbb{S}^i_{(o)},
\end{split}
\end{equation}
where $\textbf{P}_i$ is the 6D pose of the $i$-th object in the world frame, $\mathbb{S}^i_{(o)}$ is a set of suction poses in the object frame and $\mathbb{S}^i_{(w)}$ contains the corresponding poses in the world frame.

The seal scores can be directly obtained from each single objects since they only depend on local surface geometry while the wrench scores should be annotated per-scene since the wrenches depend on object poses. We also check collision to avoid invalid suctions. Specifically, we model the suction end-effector as a cylinder and check if there are object points in it. After the above three steps, we get dense suction annotations for each scene. The key parts of annotation is illustrated in Fig.~\ref{fig:annotation}. 

The ratios of each score interval for both Seal Score and Wrench Score are listed in Table~\ref{tab:anno_distribution}. Seal scores are mostly distributed in the ranges of $[0.0, 0.2)$ and $[0.8, 1.0]$ since it is usually quite clear whether it can form a seal or not while wrench scores are distributed more evenly since suction points are sampled uniformly from the 3D space.


\subsection{Evaluation Metrics}
\label{sec:evaluation}
\paragraph{Dataset Split} 

We use the same split as \cite{graspnet}. Among all 190 scenes, there are 100 for training and 90 for testing. The testing set can be divided into 3 categories: 30 scenes with seen objects, 30 with unseen but similar objects and 30 for novel and adversarial objects. As we can see in Table~\ref{tab:anno_distribution}, the Test-Seen and Test-Similar set share similar distribution with the Train set.  The Test Novel set is very hard, where only a small number of suctions get high scores.

\paragraph{Metrics} 

Given a suction pose, we first associate it with the target object by checking which object is closest to the suction point. Then the evaluation process is similar with annotation. After computing the seal score $S_{seal}$ and wrench score $S_{wrench}$ separately, the final score is obtained by production $S = S_{seal} \times S_{wrench}$.

Given a cluttered scene, suction pose prediction algorithms should predict multiple grasps. Following \cite{graspnet}, we use \textit{Precision@k} as our metrics, which measures the precision of top-\textit{k} ranked suctions. Since the evaluation score is continuous instead of binary, we define several score thresholds to decide whether a suction is positive, that is, if the score of a suction is larger than a threshold $s$, then it is regarded as positive under threshold $s$. AP$_s$ denotes the average of \textit{Precision@k} under the threshold $s$, where \textit{k} ranges from 1 to 50. Similar to previous works~\cite{graspnet}, AP$_s$ under different threshold $s$ is reported. We also report the overall $\textbf{AP}$ as the average of AP$_s$ with $s$ ranging from 0.2 to 0.8, with an interval $\delta s = 0.2$.

To prevent hacking on testset by predicting many similar suction poses and encourage diverse suction poses, we run Non-Maximal Suppression (NMS)~\cite{graspnet} before evaluation and restrict the maximal number of suctions on a single object to 10.

Since we usually perform the best suction candidate given any scenes, we also report AP and AP$_s$ of the top-1 suction pose (which only calculates \textit{Precision@1}), denoted as AP-top1 and AP$_s$-top1.

\subsection{Summary}

As mentioned before, real-world data and dense annotations contradict to each other to some extent but thanks to our annotation system and data collected by \cite{graspnet}, we build a real-world dataset with much larger scale and diversity. Comparison with other publicly available suction dataset is shown in table~\ref{tab:comparison}.

As for evaluation, continuity in suction space leads to infinite number of possible suction candidates. Therefore, no matter by human annotation~\cite{zeng2018robotic} or simulation, it is unlikely to cover all possible suction predictions. To this end, following \cite{graspnet}, we do not provide pre-computed labels for test set, but directly evaluate them by computing scores with our evaluation system. The API will be made publicly available.

A robot system can basically be separated into two parts: perception part and control part. Our benchmark falls into the first category, focusing on the problem of suction pose prediction from visual input and leaving out the control part. We make our annotations irrelevant to real robotic design choices as much as we can. Although the data comes from the previous work~\cite{graspnet}, due to the different topology and working mechanism of the end-effectors, the annotation algorithm is totally different. As mentioned in Sec.~\ref{sec:intro}, suction has several advantages compared with parallel-jaw grasping but lacks enough investigation, we hope our benchmark can help researchers design their suction prediction algorithms and compare with others fairly.

\begin{table*}[ht]
	\centering
	\caption{Real-robot success rates / average scores in different score ranges. "-" means that there are not enough labeled suction candidates in the range.}
	\vspace{-0.3cm}
	\label{tab:realrobot}
	\begin{tabular}{|c|c|c|c|c|c|c|c|}
		\hline
		Object ID & $ S_{seal} \in [0, 0.2]$ & $ S_{seal} \in [0.4, 0.6]$ & $ S_{seal} \in [0.8, 1.0]$ & Object ID &  $ S_{seal} \in [0, 0.2]$ & $ S_{seal} \in [0.4, 0.6]$ & $ S_{seal} \in [0.8, 1.0]$ \\ \hline
		000 & 8\% / 0.045 & 48\% / 0.505 & 100\% / 0.980 & 031   & 0\% / 0.000  & - / - & 100\% / 0.863 \\ \hline
		005 & 4\% / 0.008 & 60\% / 0.538 & 92\% / 0.881 & 036 & 0\% / 0.029 & 40\% / 0.477 & 100\% / 0.977\\ \hline
		010 & 12\% / 0.075 & 56\% / 0.524 & 100\% / 0.964 & 046 & 4\% / 0.001 & 44\% / 0.492 & 100\% / 0.950 \\ \hline
		015 & 9.1\% / 0.068 & 52.9\% / 0.495 & 100\% / 0.945 & 063 & 8\% / 0.039 & 56\% / 0.509 & 92\% / 0.910 \\ \hline
		024 & 4\% / 0.026 & 48\% / 0.513 & 96\% / 0.950 & 065 & 4\% / 0.024 & 48\% / 0.523 & 96\% / 0.929 \\ \hline
	\end{tabular}
	\vspace{-0.3cm}
\end{table*}

\section{Method}
\label{sec:method}
In this section, we will describe our end-to-end suction pose prediction framework. We will first introduce the problem formulation and then describe our framework pipeline.

\subsection{Problem Formulation}

To let a robot perform a suction grasping, we need to tell it where to place the suction cup and how it should approach the object. Therefore, we define a suction grasping pose as:

\begin{equation}
\label{eq:suction_pose}
\textbf{S} = [\textbf{p}, \textbf{u}],
\end{equation}
where $\textbf{p}$ is the suction point, \textit{a.k.a.} the center of contact ring between suction cup and object and $\textbf{u}$ is the direction vector pointing from the suction point outside the object surface.

We define the prediction task as, given a RGB-D image of a cluttered scene from a specific viewpoint, the algorithm should predict a set of reasonable suction poses $\mathbb{S}$. Based on that, our framework is formulated as follows.

\subsection{Framework}
\label{sec:framework}
\begin{figure}[ht]
	\begin{center}
		\includegraphics[width=1.0\linewidth]{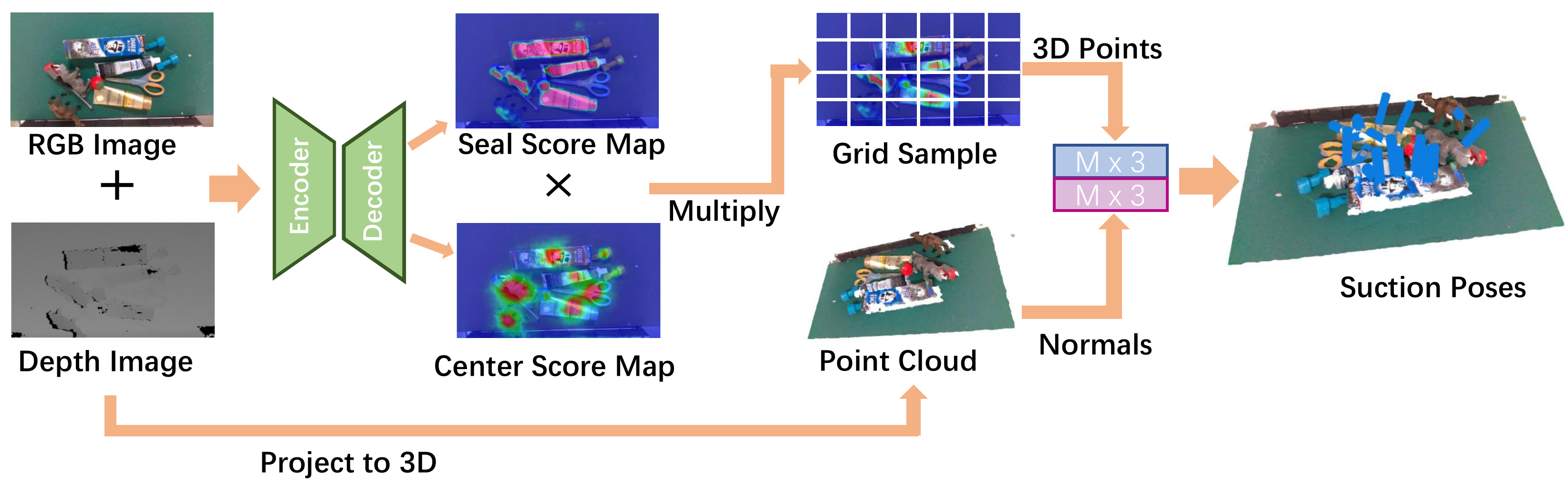} 
	\end{center}
	\vspace{-0.2cm}
	\caption{Our proposed pipeline. Given an RGB-D image, our network predicts a seal score heatmap and center score heatmap separately. The two heatmaps are then multiplied to get a final score map. We perform grid sampling on the final score and project the sampled pixels to 3D space. Suction directions are simply obtained by estimating the normals of corresponding points.}
 	\vspace{-0.3cm}
	\label{fig:framework}
\end{figure}

We assume the camera intrinsics are known so we can project a 2d pixel coordinate to its 3D point. Our pipeline is illustrated in Figure~\ref{fig:framework}, which is composed of 3 steps: 1) 2D heatmap prediction, which predicts pixel-wise suction scores; 2) sampling a fixed number of suction points; 3) suction direction estimation.

\paragraph{Heat Map Prediction} 

Instead of directly predicting the final suction scores, we decouple the heatmap into two separate heatmaps, namely seal heatmap and wrench heatmap, corresponding to seal score and wrench score mentioned in Sec.~\ref{sec:pose_anno}. Similarly, the final heatmap is obtained by multiply the two heatmaps pixel-wise.

We adopt DeepLabv3+~\cite{deeplabv3plus} for heatmap prediction because they have key designs to deal with different object sizes and to better recognize object boundary. Specifically, we use ResNet-101~\cite{resnet} as the encoder and modify the input channel of the first convolution layer from 3 to 4 (RGB to RGB-D).


Since suction poses are labeled in 3D space, to get the 2d label of seal score heatmap, we first project the 3D points onto 2D image using the camera intrinsics. To inpaint the empty pixels caused by voxel sampling, we draw a local Gaussian heatmap around each projected pixel. As for wrench heatmap label, inspired by CenterNet~\cite{centernet}, we draw a center heatmap to approximate the wrench heatmap for efficiency. The network is trained with MSE (Mean Squared Error) loss function.

\paragraph{Downsampling Suctions} 

The goal of our framework is to predict a fixed number of suction poses for each scene. However, the predicted heatmap is too dense to directly recover suction poses from pixels. Therefore, we propose a simple but effective sampling strategy. Specifically, we first divide the image to several regular grids and sample the suction with the highest predicted score within each grid. We then rank these sampled suction proposals and output the pixel coordinates of the top 1024 suctions, which can be further converted to 3D points with the camera intrinsics.

\paragraph{Suction Direction Estimation}

We simply use the surface normals of the suction points as direction. Specifically, we first create point cloud from the depth image and use Open3D~\cite{open3d} to directly estimate normals. 

\begin{figure}[ht]
	\begin{center}
		\includegraphics[width=0.8\linewidth]{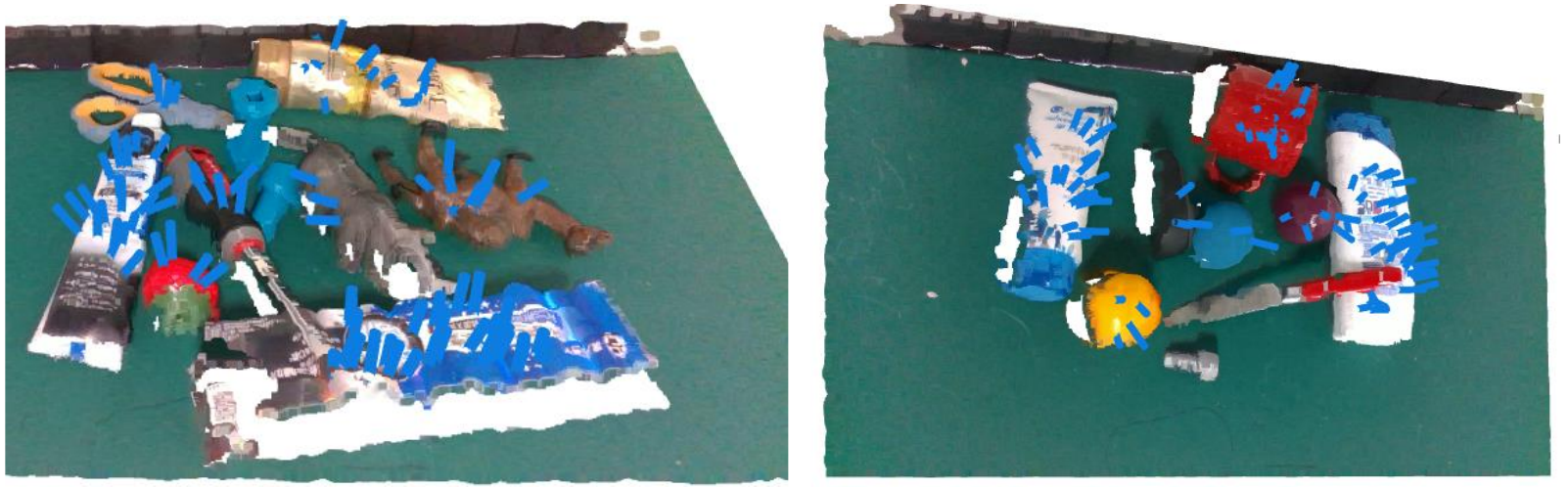} 
	\end{center}
	\vspace{-0.3cm}
	\caption{The qualitative results of our method. Scenes are constructed from RGB-D image using the camera intrinsics. Suctions are represented by blue cylinders.}
	\label{fig:qualitative}
	\vspace{-0.3cm}
\end{figure}

\begin{table*}[ht]
	\centering\footnotesize
	\caption{Evaluation for different methods. The table shows the results on data captured by Kinect/RealSense respectively.}
	\vspace{-0.3cm}
	\label{tab:method_results}
	
	\begin{tabular}{|c|c|ccc|ccc|ccc|}
		\hline
		\multirow{2}*{Metric} & \multirow{2}*{Methods} & \multicolumn{3}{c|}{Seen}           & \multicolumn{3}{c|}{Similar}       & \multicolumn{3}{c|}{Novel}          \\ 
		\cline{3-11} 
		& & \textbf{AP} & AP$_{0.8}$     & AP$_{0.4}$     & \textbf{AP} & AP$_{0.8}$     & AP$_{0.4}$     & \textbf{AP} & AP$_{0.8}$     & AP$_{0.4}$     \\ 
		\hline
		\multirow{4}*{Top-50} & Normal STD & 6.16/10.10   & 0.74/1.01 & 7.20/12.64 & 17.54/19.23   & 2.38/2.26 & \textbf{24.68}/25.95 & 1.34/3.50   & 0.12/0.23 & 1.08/4.22 \\
		& DexNet3.0~\cite{dexnet3.0} & 9.46/15.50   & 0.96/1.53 & 11.90/20.22 & 14.93/18.92   & \textbf{2.45}/2.62 & 19.05/24.51 & 1.68/2.62   & 0.04/0.35 & 1.89/5.32 \\
		& Zeng \textit{et al.}~\cite{zeng2018robotic} & 14.82/28.31   & 1.00/3.37 & 20.02/\textbf{38.58} & 14.23/26.15   & 1.61/3.35 & 18.66/34.83 & 2.85/7.79   & 0.17/\textbf{0.38} & 3.35/9.78 \\ 
		\cline{2-11} 
		& Ours & \textbf{15.94}/\textbf{28.31} & \textbf{1.25}/\textbf{3.41} & \textbf{20.89}/38.56 & \textbf{18.06}/\textbf{26.64} & 1.96/\textbf{3.42} & 24.00/\textbf{35.34} & \textbf{3.20}/\textbf{8.23} & \textbf{0.23}/0.35 & \textbf{3.78}/\textbf{10.29} \\ 
		\hline
		\multirow{4}*{Top-1} & Normal STD & 11.68/15.40   & 1.94/1.82 & 13.35/18.83 & \textbf{34.11}/31.20   & \textbf{6.51}/4.74 & \textbf{47.28}/43.46 & 2.64/4.89   & \textbf{0.28}/\textbf{0.52} & 3.72/5.96 \\
		& DexNet3.0~\cite{dexnet3.0} & 12.30/10.61   & \textbf{3.36}/1.30 & 15.03/13.91 & 20.75/14.28   & 5.39/2.40 & 26.00/17.71 & 1.09/2.81   & 0.08/0.29 & 1.17/3.15 \\
		& Zeng \textit{et al.}~\cite{zeng2018robotic} & 25.48/42.59   & 1.97/6.78 & 34.84/60.70 & 14.88/35.56   & 1.46/4.52 & 19.02/48.95 & 7.13/13.15   & 0.17/0.12 & 7.98/19.80 \\ 
		\cline{2-11} 
		& Ours & \textbf{26.44}/\textbf{43.98} & 2.90/\textbf{7.79} & \textbf{35.98}/\textbf{60.81} & 24.59/\textbf{36.59} & 2.45/\textbf{6.54} & 32.89/\textbf{49.30} & \textbf{6.66}/\textbf{14.32} & 0.13/0.38 & \textbf{8.80}/\textbf{19.95} \\ \hline
	\end{tabular}
	\vspace{-0.3cm}
\end{table*}

\section{Experiments}
\label{sec:exp}
In this section, we will first show that our ground-truth annotations align well with real-world suction through robotic experiments and then compare several representative methods with our method using the unified metrics introduced in Sec.~\ref{sec:evaluation}.

\vspace{-0.1cm}
\subsection{Annotation Evaluation}
\label{sec:realrobot}

We conduct real-robot experiments to show that our Seal Score annotations align well with the real world. The Wrench Score annotations are omitted because we make it a relative score to remove the dependence on specific object weights. We paste ArUco code on the objects to get their 6D poses in our experiment. 

\begin{figure}[ht]
	\begin{center}
		\includegraphics[width=0.8\linewidth]{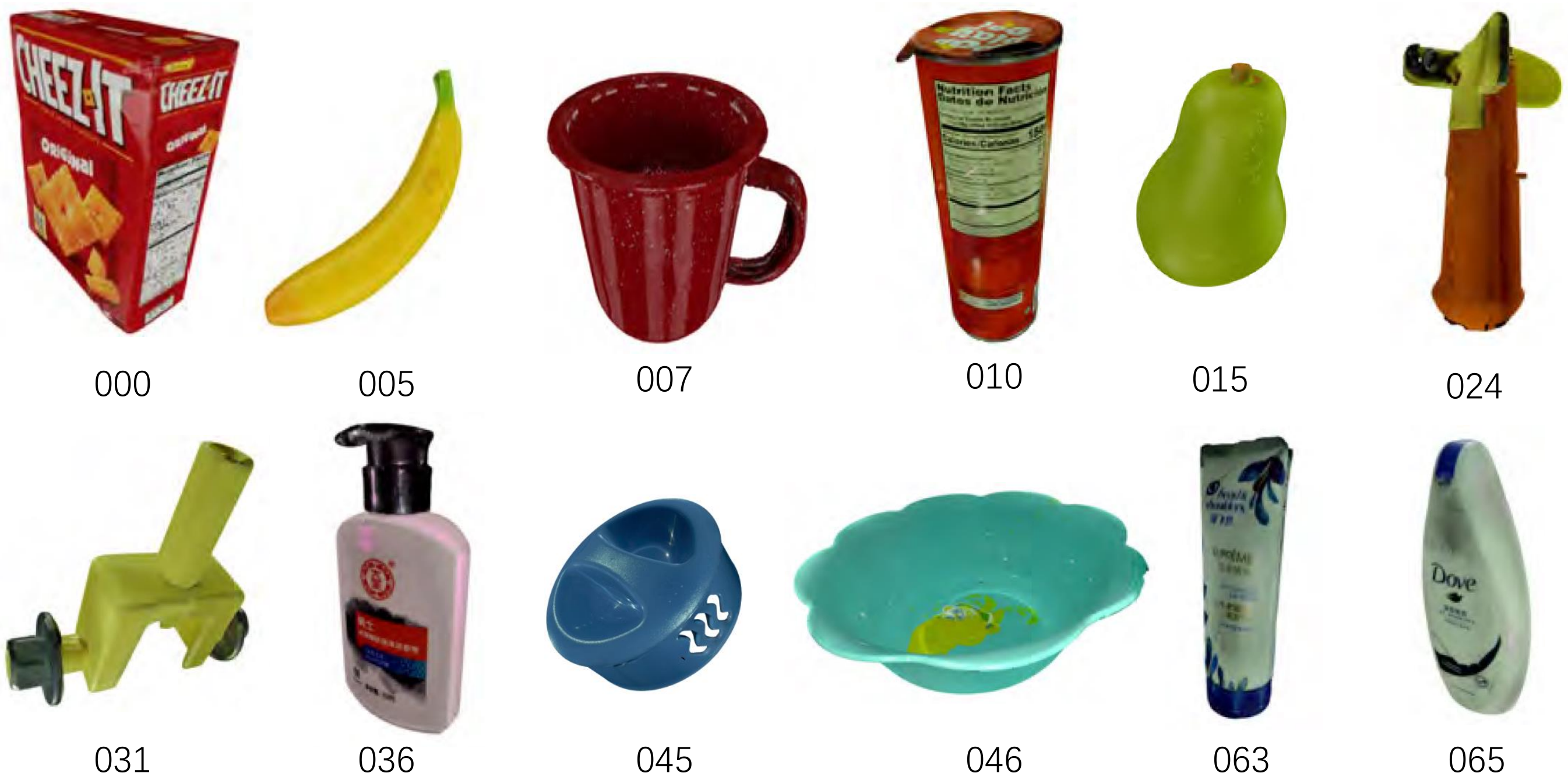} 
	\end{center}
	\vspace{-0.3cm}
	\caption{objects we use to do real-robot evaluation.}
	\vspace{-0.3cm}
	\label{fig:realrobot}
\end{figure}

For the hardware setting, we use a common UR5 robotic arm with an Intel RealSense D435 RGB-D camera. 12 objects are picked from our dataset (shown in Fig.~\ref{fig:realrobot}). To verify the consistency of our annotation and the real-world suction success rate, we choose the annotated suction poses from three representative score ranges: [0,0.2], [0.4,0.6], [0.8,1.0] and conduct real-world suction. For each score range, 25 suction poses are picked randomly. The suction success rates and the average annotation scores of executed suction poses are shown in Table~\ref{tab:realrobot}. As we can see, the success rates align well with the scores.

\vspace{-0.1cm}
\subsection{Baseline Introduction}
\label{sec:baseline}
We evaluate three baseline methods listed below:

\paragraph{Normal STD}
This method estimates the corresponding normal for each pixel and compute the standard deviation (STD) $\sigma$ within a local patch. The $\sigma$ is normalized to $[0, 1]$ by dividing the maximal standard deviation value in the image. The final score for each pixel is computed as $S = 1 - \sigma$. A bounding box indicating the location of the objects is used as an additional input otherwise most predicted suction points will lie on the flat table.

\paragraph{DexNet 3.0} 
Mahler \textit{et al.}~\cite{dexnet3.0} trained a network named GQCNN to predict a score given the suction pose and the depth image centered at the suction point. To get suction poses, they first uniformly sample initial suction poses, predict their scores and finally post process the suctions with Cross Entropy Method.
We directly evaluate their pretrained model on our benchmark.

\paragraph{Fully Convolutional Network}
Zeng \textit{et al.}~\cite{zeng2018robotic} propose to use a fully convolutional network to predict heatmap given a RGB-D image. Specifically, they use two branches of ResNet-101~\cite{resnet} to learn from RGB image and depth image separately. The features from two branches are then fused to predict suction score. Finally, the predicted scores are upsampled bilinearly to original resolution. We use the same training data, loss function, downsampling and normal estimation steps as ours, mentioned in Sec.~\ref{sec:framework}, to modify the network to our setting.
\vspace{-0.1cm}
\subsection{Results and Analysis}

The results are listed in Table~\ref{tab:method_results}. We provide some analysis below.

\subsubsection{Performances of Different Algorithms} 
\label{sec:performance_different_alg}
As we can see, our method performs the best among all the methods even without additional object bounding box as input. 
DexNet 3.0~\cite{dexnet3.0} performs worse than \cite{zeng2018robotic} and our method. There can be two reasons. The first one is due to domain gap as shown in Section~\ref{sec:domain_gap_analysis}. The second is that DexNet 3.0 was trained with single objects so it had difficulty transferring to multi-object scenes. The evidence of the difficulty can be found in Figure 3.C of DexNet 4.0~\cite{dexnet4.0}, the network trained with 100 objects and 2.5k heaps has a higher reliability compared with the one trained with 1.6k objects and 100 heaps, especially in level 2.

\subsubsection{Performances on Different Splits}

As we can see, performances of different algorithms on the Test-Similar set is generally the highest. The reason is that this set contains objects easier to perform suction on, just as indicated in Table~\ref{tab:anno_distribution}.
Though simpler, the set is still suitable to evaluate algorithms' ability of generalization, which is indicated by the relative performance difference between the Normal STD method and our neural network. Since the Normal STD method does not have the problem of generalization, less advantage of our method indicates the need of generalization.
The performances on the Test Novel set are all quite low. The reason is that the set contains new and adversarial objects with few proper suction points, as shown in Table~\ref{tab:anno_distribution}

\subsubsection{Top-1 performance}
We can see from~\ref{tab:method_results} that our method also outperforms previous methods under the Top-1 metric. An exception is that our network fails to outperform the traditional Normal STD method on the Similar split with Kinect. The reason is that Kinect camera has a wide view so the objects look small in the images. with an additional bounding box indicating where the objects are, the traditional method does not face the problem. In practice, we can eliminate such problem by cropping the images.

\vspace{-0.1cm}

\subsection{Ablation Study}

\subsubsection{Domain Gap Analysis} 
\label{sec:domain_gap_analysis}
We analyze the domain gap of both RGB images and depth images. To show the domain gap of RGB images, since our network is trained with real RGB-D images, we further test it on the 
synthetic RGB with real depth images
and see how much the performance will decrease, which can give us the same insight with training on synthetic data and testing on real data. To show the domain gap of depth images, we compare the performance difference of DexNet 3.0~\cite{dexnet3.0} since it only takes depth images as input.

Table~\ref{tab:abla_domain_gap} shows the results on the similar split. There do exist domain gaps with regard to the two different inputs, especially for RGB images. The results of depth images are relatively small but the gap will increase if we only compare the top 1 suction poses. 

\begin{figure}[ht]
	\begin{center}
		\includegraphics[width=0.8\linewidth]{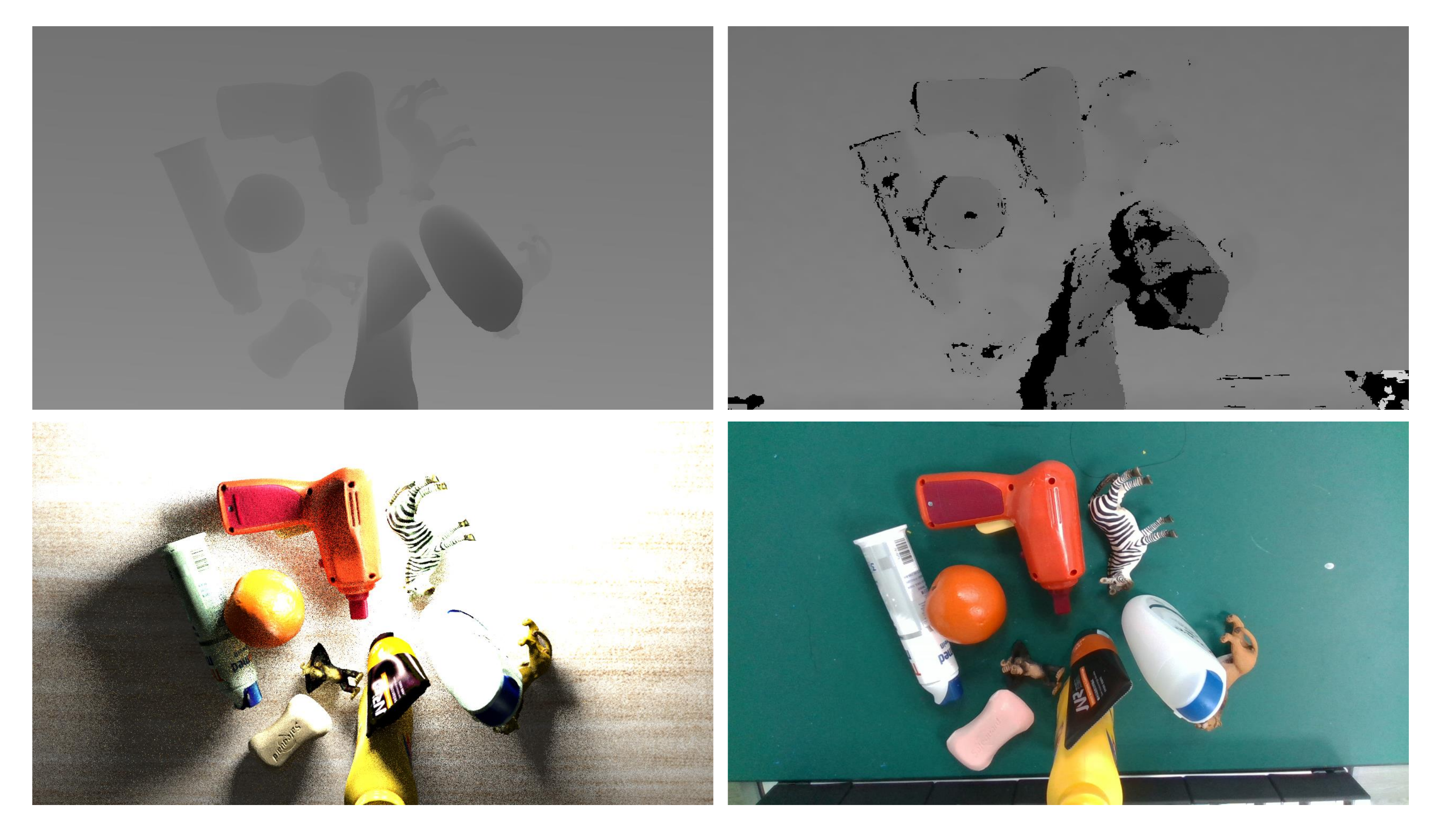}
	\end{center}
	\vspace{-0.3cm}
	\caption{Examples of both synthetic (left column) and real (right column) depth images (top row) and RGB images (bottom row).}
	\vspace{-0.3cm}
	\label{fig:domain_gap}
\end{figure}

Examples of the depth images and RGB images can be found in Figure~\ref{fig:domain_gap}.

\begin{table}[ht]
	\centering
	\caption{Ablation study of the domain gap}
	\vspace{-0.2cm}
	\label{tab:abla_domain_gap}
	\resizebox{\linewidth}{!}{
	\begin{tabular}{|c|c|c|c|c|c|}
		\hline
		Network                     & Data            & Num Suctions & AP    & AP$_{0.8}$ & AP$_{0.4}$ \\ \hline
		\multirow{4}{*}{Ours}       & Synthetic RGB   & Top 50     & 7.75  & 1.39       & 10.07      \\ \cline{2-6} 
		& Real RGB        & Top 50     & 26.64 & 3.42       & 35.33      \\ \cline{2-6}
		& Synthetic RGB   & Top 1     & 3.33  & 0.00       & 0.00      \\ \cline{2-6} 
		& Real RGB        & Top 1     & 36.59 & 6.54       & 49.30      \\\hline
		\multirow{4}{*}{DexNet 3.0~\cite{dexnet3.0}} & Synthetic depth & Top 50     & 20.61 & 2.64       & 26.36      \\ \cline{2-6} 
		& Real depth      & Top 50     & 18.92 & 2.62       & 24.51      \\ \cline{2-6} 
		& Synthetic depth & Top 1      & 17.79 & 2.99       & 22.01      \\ \cline{2-6} 
		& Real depth      & Top 1      & 14.28 & 2.40       & 17.71      \\ \hline
	\end{tabular}
	}
	\vspace{-0.3cm}
\end{table}

\subsubsection{Different data scale} To show the necessity of the large data scale, we test the results of training with different data scale: $1/10$, $1/3$ and full dataset. The results can be found in Table~\ref{tab:scale_abla}. The performance increases significantly with the increase of data scale.

\begin{table}[ht]
	\centering
	\caption{Results trained with different dataset scales.}
	\vspace{-0.2cm}
	\label{tab:scale_abla}
	\begin{tabular}{|c|c|c|l|}
		\hline
		scale  & AP    & $AP_{0.8}$ & $AP_{0.4}$ \\ \hline
		$1/10$ & 3.71  & 0.30       & 4.67       \\ \hline
		$1/3$  & 23.93 & 2.94       & 31.53      \\ \hline
		full   & 26.64 & 3.42       & 35.33      \\ \hline
	\end{tabular}
\end{table}
\subsubsection{Effect of RGB} In this section, we study the effect of RGB input. The results are shown in Table.~\ref{tab:abla_pure_depth}. As we can see, model with extra RGB input perform much better on the seen split, especially for the top-1 predictions. Also, the results on the novel split indicates that RGB data can help make better predictions on hard objects. For the similar set, the model with only depth input performs better on realsense camera. We suppose that's because depth has better generalization ability compared with RGB, just as shown in Table~\ref{tab:abla_domain_gap}.

\begin{table}[ht]
	\centering
	\caption{Ablation study of RGB Input on kinect/realsense camera.}
	\vspace{-0.2cm}
	\label{tab:abla_pure_depth}
	\resizebox{\linewidth}{!}{
	\begin{tabular}{|c|c|c|c|c|}
		\hline
		Num Suctions                     & Input            &  Seen    & Similar & Novel \\ \hline
		\multirow{2}{*}{Top 50}       & RGB-D       & \textbf{15.29}/\textbf{28.32}  & \textbf{18.06}/26.64       & \textbf{3.33}/\textbf{8.23}      \\ \cline{2-5} 
		& Only Depth     & 13.84/22.89 & 17.57/\textbf{28.83}       & 3.23/7.31      \\ \hline
		\multirow{2}{*}{Top 1} & RGB-D      & \textbf{26.44}/\textbf{43.98} & \textbf{24.59}/36.59       & \textbf{6.66}/\textbf{14.32}      \\ \cline{2-5} 
		& Only Depth     & 19.52/30.54 & 18.26/\textbf{39.39}       & 4.69/10.47      \\ \hline
	\end{tabular}
	}
	\vspace{-0.3cm}
\end{table}

\subsection{Real-robot Experiments}
To show the effectiveness of our overall system, we deploy our network and DexNet 3.0~\cite{dexnet3.0} separately on a UR5 robot arm and compare their performances. To get a comprensive comparison, we adopt three different metrics. The first is the ratio of the number of successful grasps to the number total grasps, denoted as $R_{grasp}$; the second is the ratio of the number of successfully cleared objects to the number of total objects, denoted as $R_{object}$; the last one is the ratio of the number of successfully cleared objects to the number of total grasps on those objects (the number of total grasps minus the grasps on the failed objects), denoted as $R_{mixture}$. We judge if the experiment terminates by observing if the robot fails to grasp any object in consecutive 3 times. As shown in Table~\ref{tab:heap_clearing}, our network outperforms DexNet3.0~\cite{dexnet3.0} in terms of all the three metrics.

\begin{table}[ht]
	\centering
	\caption{Performance comparison between DexNet3.0~\cite{dexnet3.0} and our network in terms of three different metrics described in the above paragraph.}
	\vspace{-0.2cm}
	\label{tab:heap_clearing}
	\begin{tabular}{|c|c|c|}
		\hline
		Metrics       & DexNet 3.0~\cite{dexnet3.0}        	& Ours          \\ \hline
		$R_{grasp}$   & $16/28=57.14\%$ & $25/31=80.65\%$ \\ \hline
		$R_{object}$  & $16/20=80\%$    & $25/25=100\%$   \\ \hline
		$R_{mixture}$ & $16/21=76.19\%$ & $25/31=80.65\%$ \\ \hline
	\end{tabular}
	\vspace{-0.5cm}
\end{table}

\section{Conclusion}

In this paper, we provide a large-scale dataset annotated on real-world data of cluttered scenes. To get grid of the need of intensive human labor and expertise, we design a physical model to analytically annotate large numbers of suction poses and adapt a two-step methodology to generate scene annotations from single object annotations. The accuracy of analytical annotations is proved by real-robot experiments. An unified evaluation system is further provided for fair comparison between algorithms. Meanwhile, we propose a pipeline to predict suction poses from an RGB-D image 
and provide detailed analysis. 
We hope our work can benefit related research.




\bibliographystyle{IEEEtran}
\bibliography{IEEEexample}

\end{document}